%% file: main.tex
\documentclass[10pt,twocolumn,letterpaper]{article}

\usepackage[pagenumbers]{iccv} % To force page numbers, e.g. for an arXiv version


\definecolor{iccvblue}{rgb}{0.21,0.49,0.74}
\usepackage[pagebackref,breaklinks,colorlinks,allcolors=iccvblue]{hyperref}

\def\paperID{*****} % *** Enter the Paper ID here
\def\confName{ICCV}
\def\confYear{2025}

\title{Aria Gen 2 Pilot Dataset}

\author{Chen Kong, James Fort, Aria Kang, Jonathan Wittmer, Simon Green, Tianwei Shen, \\
Yipu Zhao, Cheng Peng, Gustavo Solaira, Andrew Berkovich, Nikhil Raina, \\
Vijay Baiyya, Evgeniy Oleinik, Eric Huang, Fan Zhang, Julian Straub, \\
Mark Schwesinger, Luis Pesqueira, Xiaqing Pan, Jakob Julian Engel, Carl Ren,\\ 
Mingfei Yan, Richard Newcombe\\
}

\begin{document}
\maketitle
\input{sec/0_abstract}    
\input{sec/1_intro}
\input{sec/2_description}
\input{sec/3_content}
\input{sec/4_tools}
\input{sec/5_end}
{
    \small
    \bibliographystyle{ieeenat_fullname}
    \bibliography{main}
}

\end{document}

%% file: sec/0_abstract.tex
\begin{abstract}
The Aria Gen 2 Pilot Dataset (A2PD) is an egocentric multimodal open dataset captured using the state-of-the-art Aria Gen 2 glasses~\cite{AriaGen2Device}. To facilitate timely access, A2PD is released incrementally with ongoing dataset enhancements. The initial release features Dia'ane, our primary subject, who records her daily activities alongside friends, each equipped with Aria Gen 2 glasses. It encompasses five primary scenarios: cleaning, cooking, eating, playing, and outdoor walking. In each of the scenarios, we provide comprehensive raw sensor data and output data from various machine perception algorithms. These data illustrate the device's ability to perceive the wearer, the surrounding environment, and interactions between the wearer and the environment, while maintaining robust performance across diverse users and conditions. The A2PD is publicly available at \href{https://www.projectaria.com/datasets/gen2pilot/}{projectaria.com}, with open-source tools and usage examples provided in Project Aria Tools.
\end{abstract}

%% file: sec/1_intro.tex
\section{Introduction}
\label{sec:intro}
The goal of Project Aria is to enable researchers across the world to advance the state of the art in machine perception, contextual AI, and robotics through access to cutting-edge research hardware and open source datasets, models, and tooling. The foundation for this ecosystem was established with the release of Aria Gen 1 in 2020, which has had a significant impact on the research community. Since its debut, Aria Gen 1 has become the world’s most widely adopted device for egocentric research, with over 290 academic and industrial partners operating more than 1,000 devices across 27 countries, recording over 8,000 hours of data. Meta has released a suite of open-sourced datasets created using Project Aria that address the fundamental problems of understanding people within their environments, modeling the environments themselves, and capturing how humans interact with and alter them. These datasets have reached thousands of external users, with over 5,000 unique downloads. Additionally, Aria partners have developed groundbreaking benchmarking datasets that have shaped how AI systems are evaluated. There are now over 600 citations of the Aria device and its associated datasets across research areas spanning computer vision, contextual AI, robotics, augmented reality, human–computer interaction, and assistive technologies.

Now we will review a subset of Aria-based datasets from Meta and Aria partners that illustrates the typical research focus areas with the device to date.

{\bf Aria Everyday Activities (AEA)}~\cite{lv2024aria}: AEA democratizes access to naturalistic human behavior data, offering 143 sequences spanning over seven hours of daily activities. These recordings capture authentic human interactions in shared spatial contexts, providing time-synchronized data essential for developing contextual AI systems.

{\bf Nymeria}~\cite{ma2024nymeria}: Nymeria is the world's largest dataset of human motion in the wild, capturing diverse people engaging in diverse activities across diverse locations. It is first of its kind to record body motion using multiple egocentric multimodal devices, all accurately synchronized and localized in one single metric 3D world.

{\bf HOT3D}~\cite{banerjee2025hot3d}: HOT3D is a dataset for egocentric 3D hand and object tracking with eye gaze, featuring 3.7M+ images from 19 subjects interacting with 33 objects in varied environments. It includes multi-view RGB/monochrome images, eye gaze, point clouds, and 3D poses, recorded with Aria glasses and Quest 3 headsets, enabling advanced benchmarking for hand-object interactions.

{\bf Aria Digital Twin (ADT)}~\cite{pan2023aria}: This dataset provides 200 sequences captured in two instrumented indoor environments, featuring 398 object instances. ADT offers comprehensive ground-truth annotations, including continuous 6-degree-of-freedom poses for devices and objects, 3D eye gaze vectors, human pose annotations, instance segmentations, and depth maps, setting new benchmarks for egocentric machine perception.

{\bf Ego-Exo4D}~\cite{grauman2024ego}: Developed by the Ego4D consortium, Ego-Exo4D presents three meticulously synchronized natural language datasets paired with videos covering the topics of expert commentary, revealing nuanced skills, participant-provided narrate-and-act descriptions in a tutorial style, and one-sentence atomic action descriptions to support browsing, mining the dataset, and addressing benchmarks in video-language learning.

{\bf HD-EPIC}~\cite{perrett2025hd}: Developed by researchers at University of Bristol, HD-EPIC is a large-scale, unscripted egocentric video dataset recorded in nine home kitchens over 41 hours. It features exceptionally rich, interconnected 3D annotations—capturing recipe steps, nuanced hand actions, ingredient details, object movements, and audio events—providing authentic, contextually grounded insights into everyday kitchen activity at an unprecedented level of detail.
%TODO add citation for HD-EPIC

{\bf Egolife}~\cite{yang2025egolife}: Developed by researchers at Nanyang Technological University, the Egolife dataset captures long-term, real-world egocentric experiences, providing rich multi-modal data for studying daily life activities, social interactions, and environmental context. It supports research in activity recognition, social understanding, and context-aware AI.

{\bf EgoMimic}~\cite{kareer2025egomimic}: Developed by researchers at Georgia Tech, EgoMimic demonstrates the transformative potential of egocentric data for robotics. By leveraging just 90 minutes of Aria recordings paired with 3D hand tracking, EgoMimic achieved a 400\% improvement in robot task performance, illustrating how human embodiment data can dramatically accelerate robotic learning and reduce dependence on costly teleoperation demonstrations.

Building on the incredible momentum of the Aria Gen 1 program, Meta announced Aria Gen 2~\cite{AriaGen2Device} in February 2025, representing a substantial technological leap forward from the Aria Gen 1 device. This next-generation platform features a comprehensively upgraded sensor suite, including four computer vision cameras (vs two on Aria Gen 1) with an expanded field-of-view, enhanced RGB camera resolution, integrated contact and spatial microphones, a photoplethysmography (PPG) sensor for physiological monitoring (such as heart rate), and improved battery life. Additional advancements include ultra-low-power on-device machine perception, integrated speakers for real-time interaction, and Sub-GHz radio technology for sub-millisecond device time alignment.

To demonstrate these advanced capabilities, we now introduce the Aria Gen 2 Pilot Dataset (A2PD). This dataset takes inspiration from the format of previous successful Aria Gen 1 datasets and showcases the full potential of Aria Gen 2’s enhanced sensor suite and on-device machine perception. It provides researchers with a concrete artifact that they can interact with to deeply understand the device's additional hardware capabilities and the resulting improvements in data quality, temporal precision, contextual richness, and types of algorithms that can be run on the data. 

A2PD will be released in incremental fashion as more data is produced and additional algorithms are run on the dataset. This paper focuses on the initial release. We document the collection methodology, technical specifications, and exemplar perception algorithm results, establishing A2PD as a valuable resource for advancing multimodal egocentric perception research.

%% file: sec/2_description.tex
\section{Dataset Description}

The A2PD captures a weekend of daily activities involving four participants (a primary wearer, Dia'ane, and three co-participants), each equipped with Aria Gen 2 glasses. The recordings document a sequence of everyday scenarios: Dia'ane begins by cleaning her room and preparing a meal, followed by the group sharing lunch and playing ``Simon Says''. Later, Dia'ane and a friend take a walk outdoors. In total, the dataset comprises five distinct scenarios and twelve sequences, each approximately five minutes in duration. These sequences encompass a diverse range of behaviors, longitudinal context, complex hand-object interactions, frequent social interactions, varied conversational content, eye movement patterns such as reading, diverse human movement dynamics, and exposure to different lighting conditions across both indoor and outdoor environments.
% Show some photos? RGB? trajectory?

%% file: sec/3_content.tex
\section{Dataset Content}
The Aria Gen 2 pilot dataset comprises four primary data modalities:
\begin{enumerate}
\item raw sensor streams acquired directly from Aria Gen 2 devices.
\item real-time machine perception outputs generated on-device via embedded algorithms during data collection.
\item offline machine perception results produced by Machine Perception Services (MPS, see Section~\ref{sec:mps} for details) during post-processing.
\item outputs from additional offline perception algorithms.
\end{enumerate}
Modalities (1) and (2) are obtained natively from the device, whereas (3) and (4) are derived through offline processing. For comprehensive details regarding folder structure and file formats, please refer to the project website.

\subsection{Raw Sensor Data}
All recordings are acquired using a pre-defined profile, resulting in a comprehensive collection of high-fidelity, time-synchronized data suitable for a broad spectrum of research tasks, including multimodal learning, sensor fusion, and context-aware modeling. Each sequence includes the following raw sensor streams:

\begin{itemize}
    \item \textbf{Visual Data:}
    \begin{itemize}
        \item RGB video captured at 10~fps, with a spatial resolution of $2560\times1920$ pixels and auto-exposure.
        \item Four computer vision (CV) video streams at 30~fps, each with $512\times512$ resolution and auto-exposure.
        \item Binocular eye-tracking imagery at 5~fps, with $200\times200$ pixel resolution per eye.
    \end{itemize}

    \item \textbf{Motion and Environmental Data:}
    \begin{itemize}
        \item Dual inertial measurement unit (IMU) signals sampled at 800~Hz.
        \item Magnetometer readings at 100~Hz.
        \item Barometric pressure measurements at 50~Hz.
        \item Global Positioning System (GPS) coordinates at 1~Hz.
        \item Ambient temperature readings at 1~Hz.
        \item Ambient light sensor (ALS) measurements at 9.434~Hz, with a 3200~$\mu$s exposure time.
    \end{itemize}

    \item \textbf{Audio and Physiological Data:}
    \begin{itemize}
        \item Eight-channel spatial audio, including contact microphone recordings.
        \item Photoplethysmography (PPG) signals sampled at 128~Hz.
    \end{itemize}

    \item \textbf{Connectivity Data:}
    \begin{itemize}
        \item Bluetooth and Wi-Fi signal traces.
    \end{itemize}

\end{itemize}

In scenes where multiple participants are present, we also leverage the a sub-GHz radio on the Aria Gen 2 device to achieve sub-millisecond device time alignment across all devices in the recording session. This ensures that multimodal data streams from different wearers are accurately time-aligned, providing a robust foundation for research requiring fine-grained temporal correspondence between participants. The time alignment signals are released as part of this dataset.
% TODO(chenk): Revisit the accuracy of the subGHz alignment.

\subsection{On-Device Machine Perception Results}
Machine perception algorithms are run concurrently on-device during all recordings. These algorithms are natively integrated into the Aria Gen 2 glasses and run on Meta’s energy-efficient custom coprocessor. The availability of such diverse and accurate perception data creates new opportunities for research, allowing for the development of real-time prototypes and more efficient recording without the need for offline processing. One can download our pilot dataset to assess the quality and robustness.

{\bf Visual Inertial Odometry (VIO)}
Aria Gen 2 delivers robust six degrees of freedom (6DOF) tracking within a spatial frame of reference using VIO. The VIO output is generated at 10Hz with 3-DOF position, 3-DOF linear velocity, 3-DOF orientation in quaternion form, 3-DOF angular velocity and estimated direction of gravity for the odometry frame. Additionally, Aria Gen2 also produces high-frequency VIO output at the IMU rate (800Hz), by performing IMU pre-integration in addition to the regular 10Hz VIO output.

{\bf Eye Tracking}
Aria Gen 2 features an advanced camera-based eye tracking system that tracks users' gaze. This system generates the following eye tracking outputs for each eye, at up to 90Hz: the origin and direction of the individual gaze ray, the 3-DOF position of the entrance pupil, the diameter of the pupil, and whether the eye is blinking. Additionally, the system also produces for the combined gaze estimated from both eyes: the origin and direction of the combined gaze ray, vergence depth of the combined gaze, and distance between the left/right eye pupils (interpupillary distance, IPD).

{\bf Hand Tracking}
Aria Gen 2 also features a hand detection and tracking solution that tracks the wearer’s hands in 3D space. The hand tracking pipeline generates at 30Hz for each hand (left and right): 3-DOF position of the wrist, 3-DOF rotation of the wrist, and 3-DOF positions of the 21 finger joint landmarks. Across the entire dataset, 73,616 left hands and 67,893 right hands are detected, covering a diverse range of hand poses.

\subsection{Machine Perception Services Results}
\label{sec:mps}
All recordings are processed offline by our popular Machine Perception Services (MPS). Recall that MPS is a cloud service where research partners with access to the Aria Research Kit can upload their \href{https://facebookresearch.github.io/projectaria_tools/gen2/technical-specs/vrs/data-format}{VRS} and request to process them by a set of proprietary machine perception algorithms, designed for Project Aria glasses. The MPS results of each sequence can be downloaded as part of the dataset.

{\bf MPS SLAM} is applied to all collected sequences. Single Sequence Trajectory (SST) is used for cleaning and cooking sequences, while Multi-Sequence Trajectory (MST) is used for eating, playing and outdoor walking sequences, ensuring all participants share a common SLAM coordinate frame. Both SST and MST provide accurate 6DOF poses, semi-dense point clouds, and online calibration of SLAM cameras and IMUs. Note that RGB camera calibration is inherited from factory settings without further optimization.

{\bf MPS Hand Tracking} is applied to all sequences, both indoor and outdoor. The hand tracking results include 3-DOF positions of 21 landmarks per hand, consistent with Gen 1. In total, 80,295 left hand and 79,161 right hand poses are provided, covering a diverse range of hand poses, particularly in cooking scenarios. Note that the MPS hand tracking is an offline algorithm that requires more compute than the on-device hand tracking, and produces hand results with higher precision and recall.
% TODO: add more statistics with accurate numbers.

\subsection{Additional Perception Algorithms Results}
Beyond SLAM, hand tracking and eye tracking, we apply a suite of additional perception algorithms to further process the collected data. Specifically, we run directional Automatic Speech Recognition (ASR)~\cite{lin2023directional}, heart rate estimation, hand-object interaction recognition and depth estimation using Foundation Stereo~\cite{wen2025foundationstereo} on all recordings, and 3D object detection via Egocentric Voxel Lifting (EVL)~\cite{straub2024efm3d} on all indoor recordings. The results of all these algorithms are released as part of the dataset.

\subsubsection{Diarization}
The Aria Gen 2 device is capable of capturing both the wearer’s voice and the voices of others interacting with the wearer. To demonstrate this capability, we apply directional ASR~\cite{lin2023directional} to all sequences. The directional ASR algorithm distinguishes between self and others, and provides accurate start and end timestamps for each utterance. Across the entire dataset, 1644 utterances are recognized including 752 from SELF and 892 from OTHERS where the longest utterance contains 22 words. Representative examples are shown below:
\begin{quotation}
\noindent
   OTHER: What have you been up today?\\
   SELF:  You know, before you guys came over, the house was a mess.\\
   OTHER: Yeah.\\
   SELF:  So, i had to do some quick cleaning.\\
   SELF:  I went to the grocery store, that's where i got all these ingredients.\\
   SELF:  But i forgot the mayo, so it might be a bit dry.\\
   OTHER: Uh, it's okay.
\end{quotation}
Note that these transcripts are generated automatically by the algorithm and have not been manually validated; therefore, they should not be considered as ground truth.

\begin{figure}[b]
  \centering
   \includegraphics[width=\linewidth]{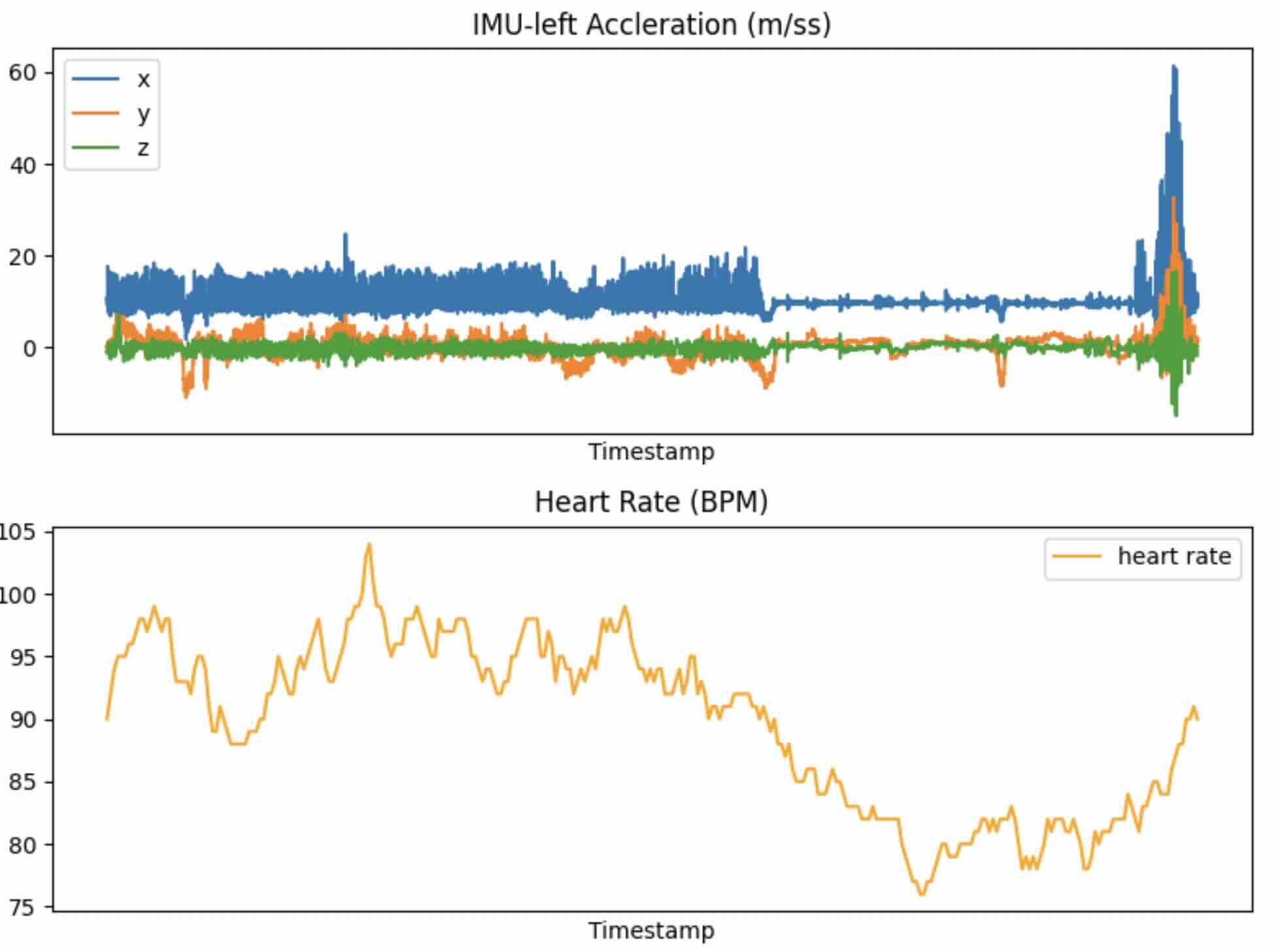}
   \caption{Plot of estimated heart rate together with IMU signals from the walking sequence. One can observe that the heart rate stays elevated during the walk, and then returns to the normal resting level afterward.}
   \label{fig:heart_rate}
\end{figure}

\subsubsection{Heart rate estimation}
The Aria Gen 2 device is equipped with photoplethysmography (PPG) sensors, enabling continuous measurement of the wearer’s heart rate. We apply our heart rate estimation algorithm to all recorded sequences and include the resulting data in the released dataset. The algorithm provides coverage for over $95\%$ of the recording duration, excluding only the initialization phase at the start and the termination phase at the end of each sequence. Although ground truth heart rate measurements are not available for this release, the estimated heart rate reliably reflects the wearer’s physical activity, with observable peaks following periods of running or jumping and lower values during rest. Validation of heart rate accuracy against chest strap sensors will be reported in a separate study. Figure~\ref{fig:heart_rate} presents an example of estimated heart rate alongside IMU signals, illustrating the correspondence between physiological and activity data.

\begin{figure}[t]
  \centering
   \includegraphics[width=\linewidth]{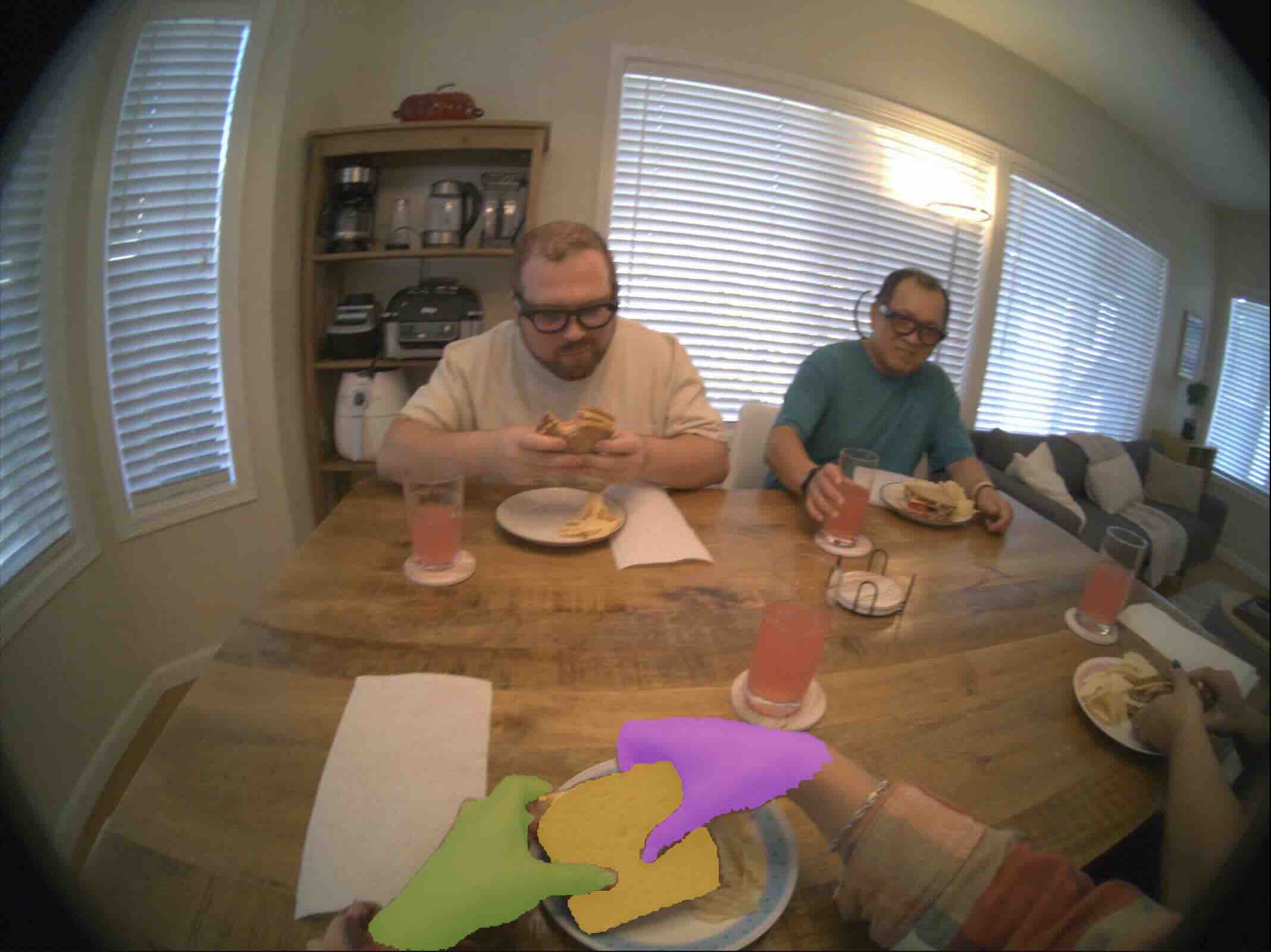}\\
   \includegraphics[width=\linewidth]{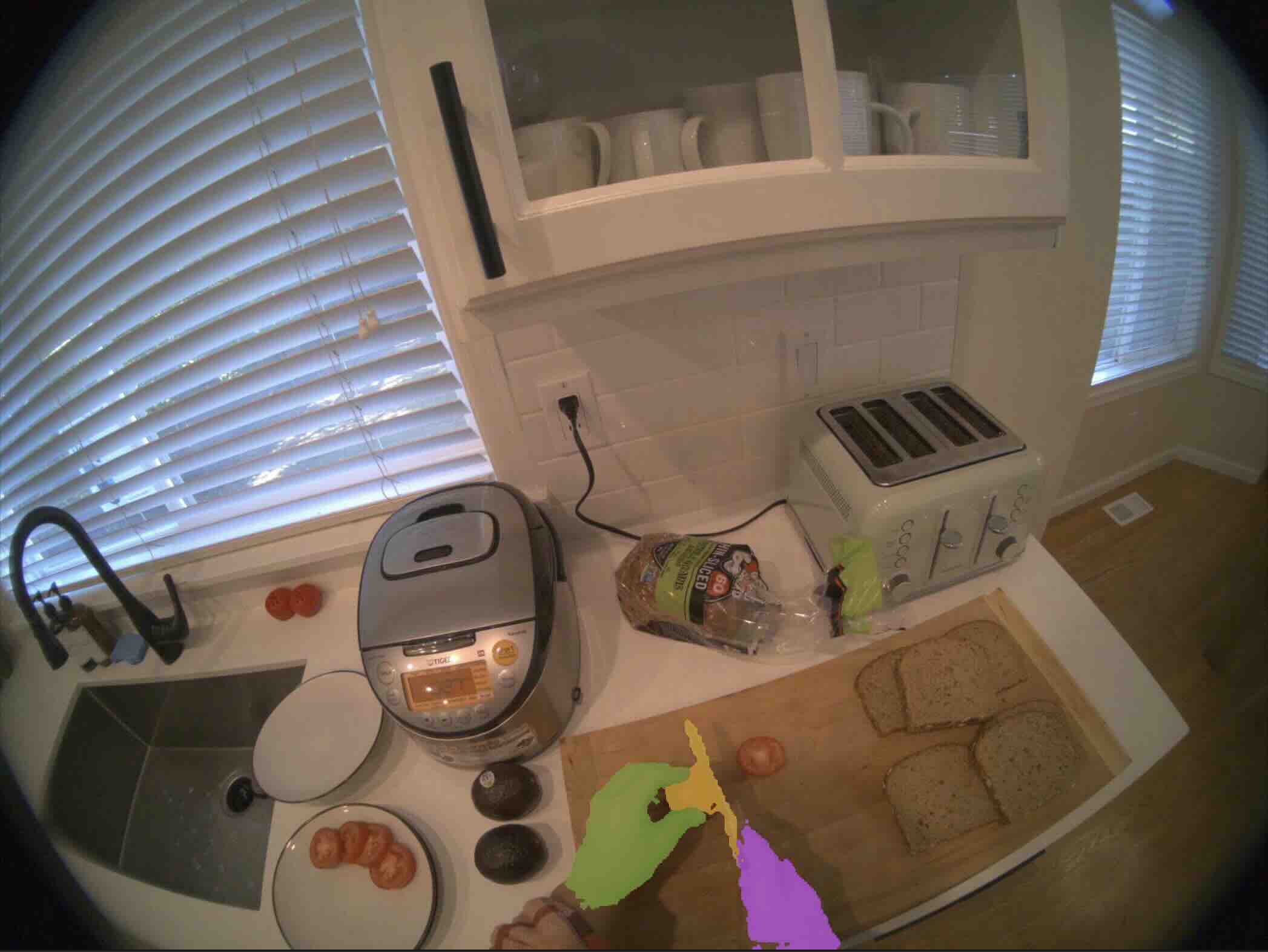}
   \caption{Example of detecting hand object interaction by segmenting hands (left hand in green, right hand in magenta) and the interacted objects (yellow).}
   \label{fig:midas}
\end{figure}

\subsubsection{Hand Object Interaction}
The Aria Gen 2 device is capable not only of estimating hand poses, but also of recognizing interactions between hands and objects. To demonstrate this capability, we trained a Mask2Former~\cite{cheng2021mask2former} using an annotated Aria dataset to detect hand-object interaction. We apply the model to the RGB stream of all sequences in the dataset, generating segmentation masks for the left hand, right hand, and interacted objects. From the RGB stream of the entire dataset, we segment a total of 15,925 left hands, 17,020 right hands, and 5,804 objects. Figure~\ref{fig:midas} presents representative results, illustrating accurate segmentation of hands and objects, as well as effective detection of their interactions.

\subsubsection{3D Object Detection}
\begin{figure}[b]
  \centering
   \includegraphics[width=\linewidth]{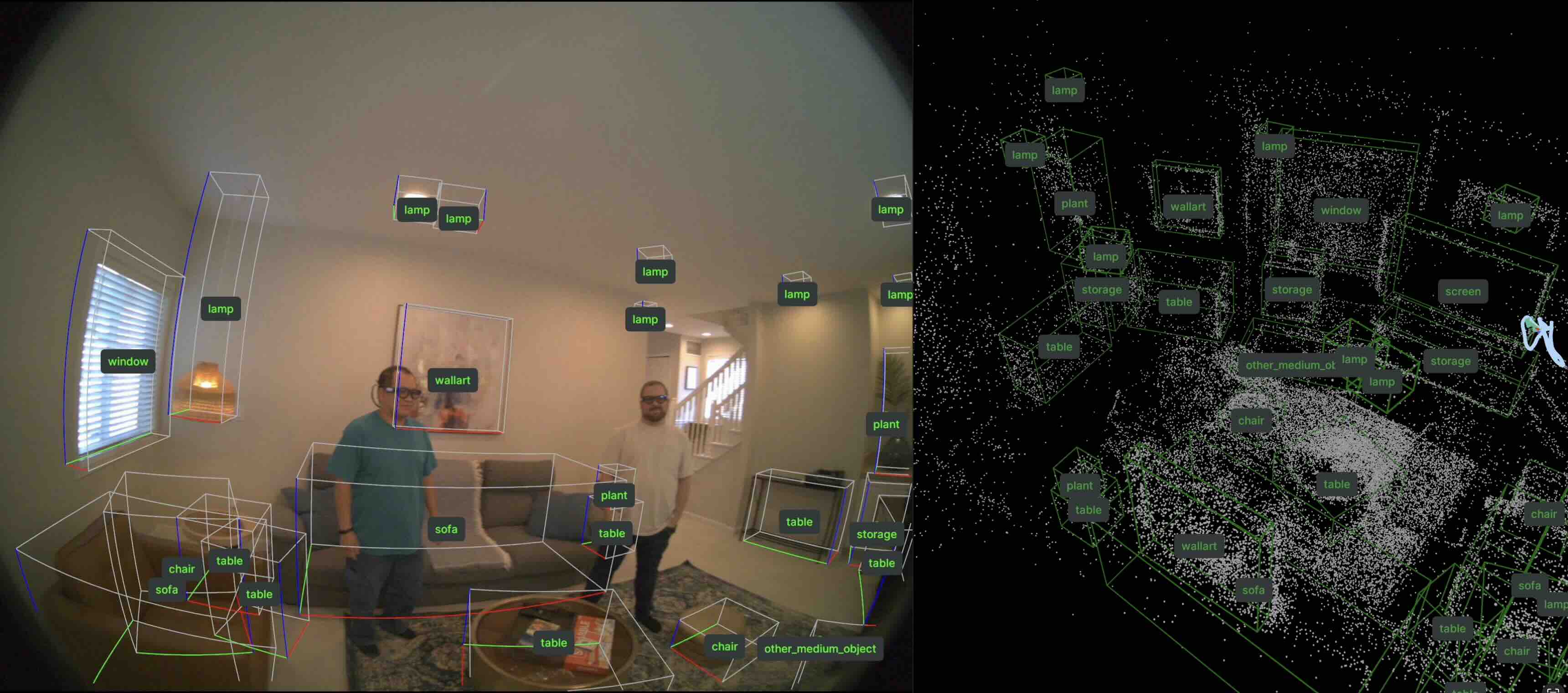}
   \caption{Example of 3D bounding boxes estimated by EVL~\cite{straub2024efm3d} projected back on RGB images and visualized together with semi-dense point clouds from MPS.}
   \label{fig:evl}
\end{figure}

The Aria Gen 2 device features a wide field-of-view multi-camera system, comprising one RGB camera and four computer vision (CV) cameras. This configuration enables comprehensive perception of the surrounding environment. To demonstrate this capability, we apply the Egocentric Voxel Lifting (EVL)~\cite{straub2024efm3d} detection and tracking algorithm to all indoor sequences, detecting environmental 3D bounding boxes. Across the dataset, we identify 293 unique 3D object bounding boxes of objects that are observed a total of 1,351,248 times. We provide both the 3D bounding boxes as well as the per frame corresponding 2D bounding boxes which indicate visibility. Figure~\ref{fig:evl} presents representative RGB images with re-projected 3D bounding boxes and detected 3D bounding boxes overlaid on semi-dense point cloud generated by MPS. The figure illustrates effective detection of large objects such as tables and chairs as well as medium sized objects like lamps, and plants.

\subsubsection{Depth Estimation}
\begin{figure}[t]
  \centering
   \includegraphics[width=\linewidth]{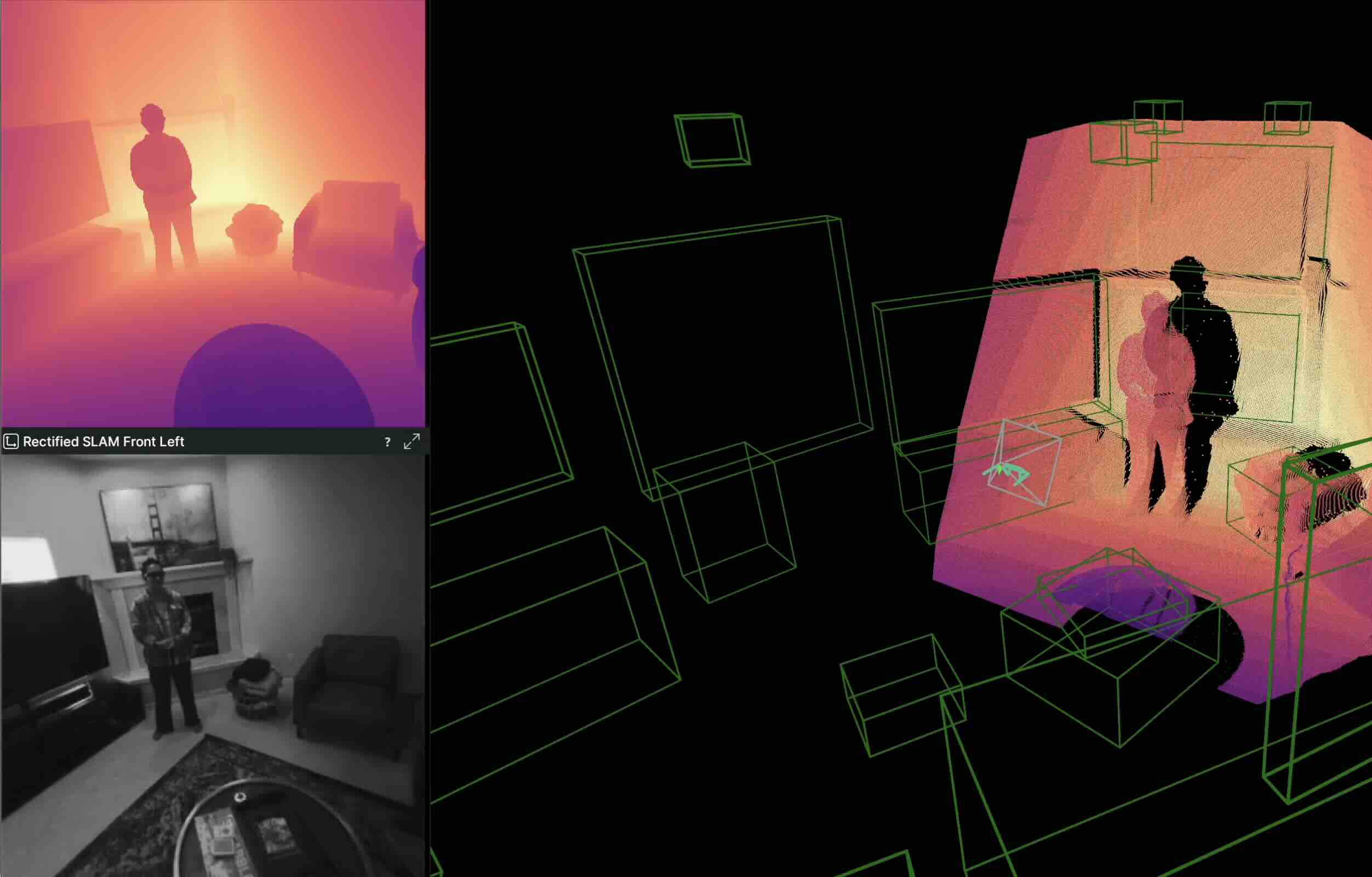}
   \caption{An example of generated depth images generated by Foundation Stereo~\cite{wen2025foundationstereo} with corresponding rectified CV images and 3D point cloud overlaid on detected 3D bounding boxes.}
   \label{fig:foundation_stereo}
\end{figure}

Accurate depth estimation has been a highly requested feature since the introduction of Aria Gen 1. With four overlapping CV cameras, Aria Gen 2 is now capable of producing reliable depth maps, effectively functioning as a precise depth capture device. To achieve this, we scan-line rectify the front left and front right CV camera images and process them using the Foundation Stereo~\cite{wen2025foundationstereo} model to generate corresponding 512 by 512 pixel depth images. The conversion from disparity to depth and world-space points makes use of Aria Gen 2's accurate online calibration.
Note that while Foundation Stereo is a state-of-the-art model and generally estimates accurate depths which closely match that of the MPS semi-dense points, it does have errors (which increase with distance), and can be inconsistent from frame to frame. We anticipate that future models tuned using Aria data will improve quality even further.
Across the entire dataset, a total of 100,415 depth images are produced. Figure~\ref{fig:foundation_stereo} presents examples of the generated depth images, along with the associated rectified CV images and 3D point clouds.

%% file: sec/4_tools.tex
\section{Dataset Tools}
\begin{figure}[t]
  \centering
   \includegraphics[width=\linewidth]{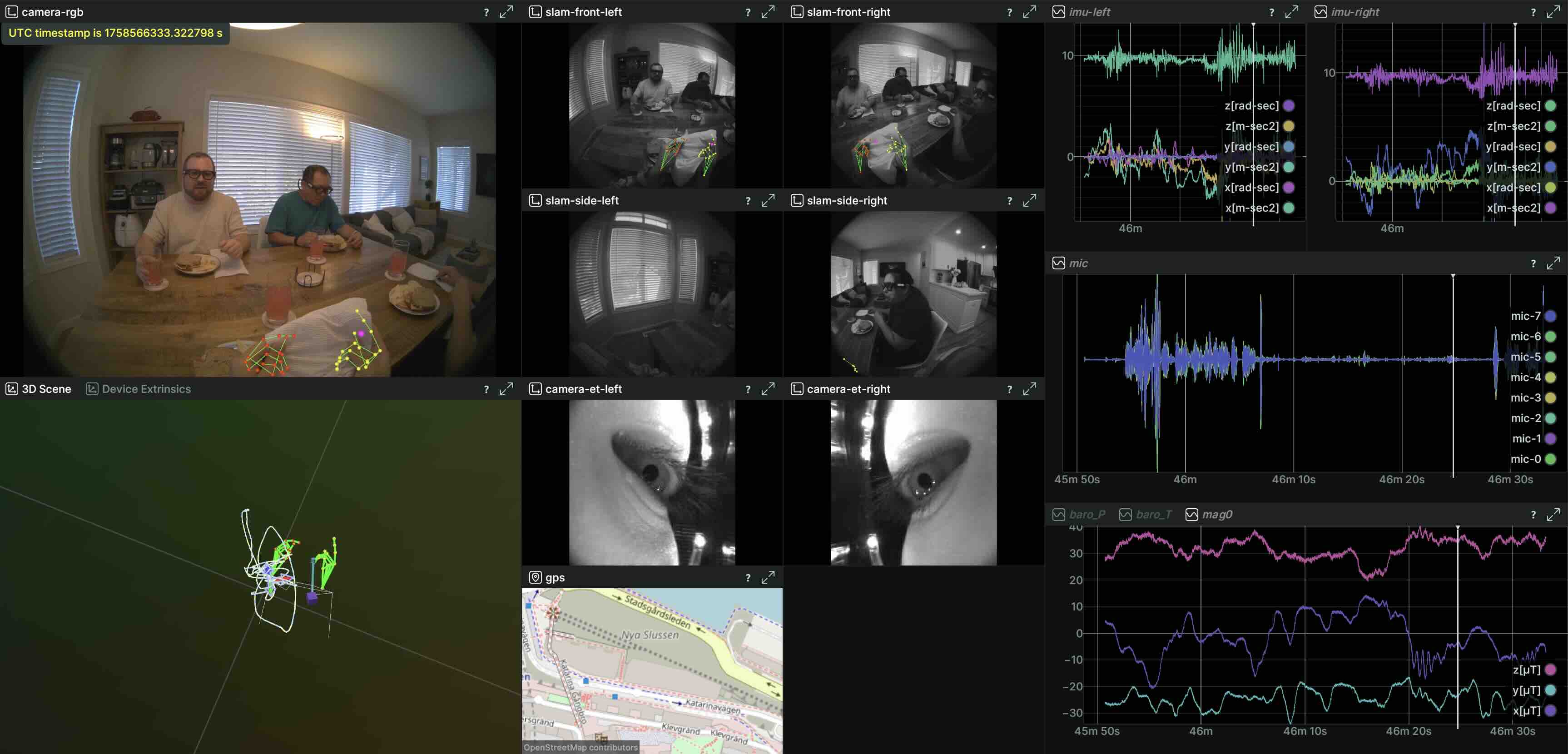}
   \includegraphics[width=\linewidth]{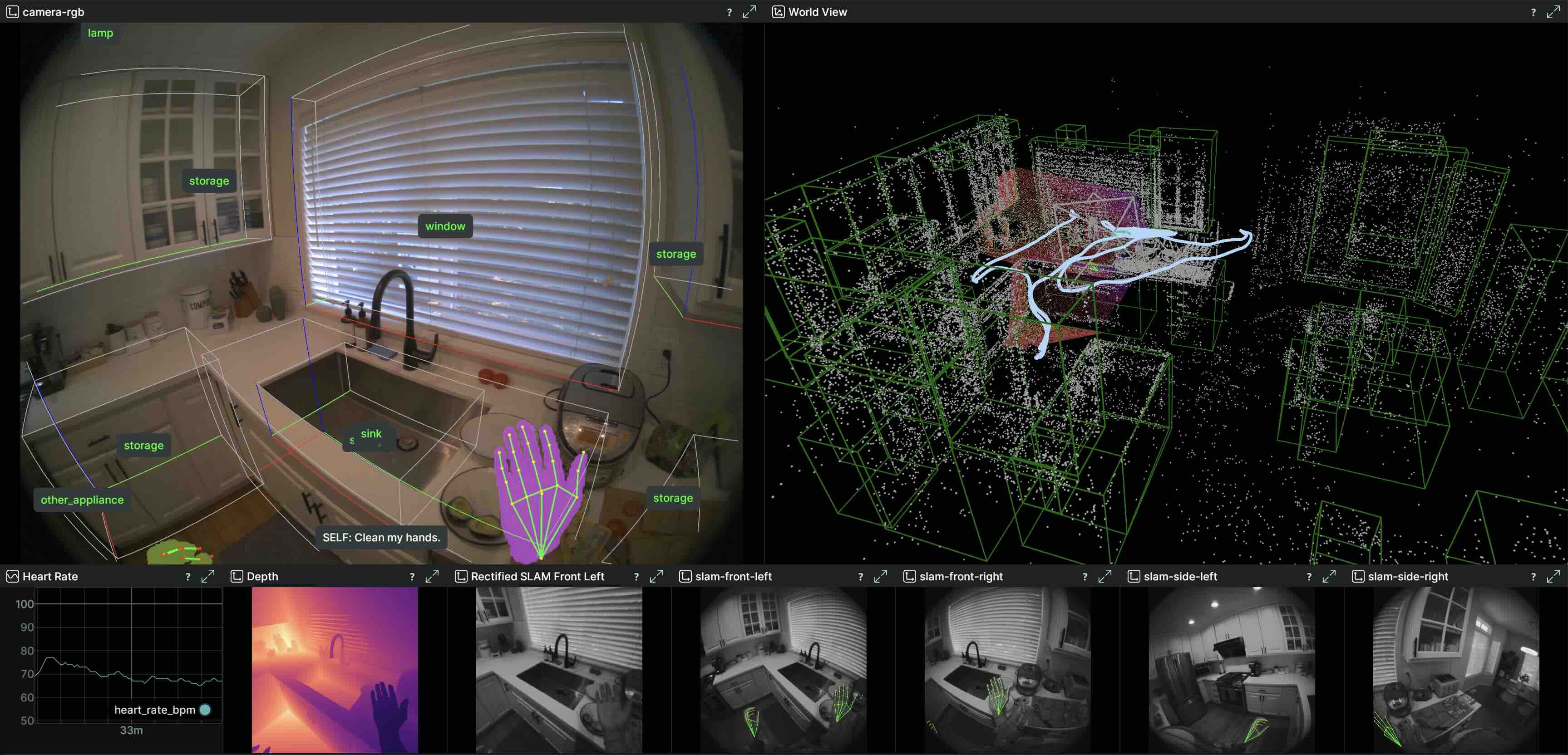}
   \caption{Screenshots of the raw sensor visualizer and machine perception visualizer.}
   \label{fig:visualizer}
\end{figure}

Together with the dataset, we are also releasing a toolkit to easily download, load and visualize the dataset. A JSON file containing download links can be retrieved from the Aria Dataset Explorer. With that file, one can run a single command to download the dataset and save it into a single location following a pre-defined folder structure. A python-based data loader has been provided to load each sequence folder. Raw sensor and MPS data are loaded into our Project Aria open source data format. We have reused the ADT~\cite{pan2023aria} data format for 3D object detection, and created and open-sourced a new data format for all of the other algorithms. Jupyter notebooks are provided to demonstrate how the data loader could be used. We also provide two visualizers: one shows each raw sensor signal and on-device machine perception data; the other one shows MPS results and offline perception algorithms results. Figure~\ref{fig:visualizer} shows a screen shot of the provided visualizers.

%% file: sec/5_end.tex
\section{Release Plan}
The Aria Gen 2 Pilot Dataset will be released incrementally and the current paper describes the first release of the dataset. Future releases will append additional data and perception algorithms to this repository. Potential algorithms are full body human motion generation, activity recognition and more. Potential future datasets may contain repeated human manipulation recordings for Robotics imitation learning, repeated robot manipulation recordings for Robotics perception, all-day long recordings for contextual AI and a complete set of detailed ground truth annotations. A2PD is hosted at \href{https://www.projectaria.com/datasets/gen2pilot/}{projectaria.com}, where the current and future releases will be made available. Users are encouraged to cite this paper for any release of this dataset.

\section*{Acknowledgements}
The authors thank Dia’ane Daniel-Richards, Lorena Esquivel, Austin Kukay, and Taylor Tran for their help with the data collection operations.

%% file: main.bbl
\begin{thebibliography}{13}
\providecommand{\natexlab}[1]{#1}
\providecommand{\url}[1]{\texttt{#1}}
\expandafter\ifx\csname urlstyle\endcsname\relax
  \providecommand{\doi}[1]{doi: #1}\else
  \providecommand{\doi}{doi: \begingroup \urlstyle{rm}\Url}\fi

\bibitem[Banerjee et~al.(2025)Banerjee, Shkodrani, Moulon, Hampali, Han, Zhang,
  Zhang, Fountain, Miller, Basol, et~al.]{banerjee2025hot3d}
Prithviraj Banerjee, Sindi Shkodrani, Pierre Moulon, Shreyas Hampali, Shangchen
  Han, Fan Zhang, Linguang Zhang, Jade Fountain, Edward Miller, Selen Basol,
  et~al.
\newblock Hot3d: Hand and object tracking in 3d from egocentric multi-view
  videos.
\newblock In \emph{Proceedings of the Computer Vision and Pattern Recognition
  Conference}, pages 7061--7071, 2025.

\bibitem[Cheng et~al.(2022)Cheng, Misra, Schwing, Kirillov, and
  Girdhar]{cheng2021mask2former}
Bowen Cheng, Ishan Misra, Alexander~G. Schwing, Alexander Kirillov, and Rohit
  Girdhar.
\newblock Masked-attention mask transformer for universal image segmentation.
\newblock In \emph{CVPR}, 2022.

\bibitem[Grauman et~al.(2024)Grauman, Westbury, Torresani, Kitani, Malik,
  Afouras, Ashutosh, Baiyya, Bansal, Boote, et~al.]{grauman2024ego}
Kristen Grauman, Andrew Westbury, Lorenzo Torresani, Kris Kitani, Jitendra
  Malik, Triantafyllos Afouras, Kumar Ashutosh, Vijay Baiyya, Siddhant Bansal,
  Bikram Boote, et~al.
\newblock Ego-exo4d: Understanding skilled human activity from first-and
  third-person perspectives.
\newblock In \emph{Proceedings of the IEEE/CVF Conference on Computer Vision
  and Pattern Recognition}, pages 19383--19400, 2024.

\bibitem[Kareer et~al.(2025)Kareer, Patel, Punamiya, Mathur, Cheng, Wang,
  Hoffman, and Xu]{kareer2025egomimic}
Simar Kareer, Dhruv Patel, Ryan Punamiya, Pranay Mathur, Shuo Cheng, Chen Wang,
  Judy Hoffman, and Danfei Xu.
\newblock Egomimic: Scaling imitation learning via egocentric video.
\newblock In \emph{2025 IEEE International Conference on Robotics and
  Automation (ICRA)}, pages 13226--13233. IEEE, 2025.

\bibitem[Lin et~al.(2023)Lin, Moritz, Xie, Kalgaonkar, Fuegen, and
  Seide]{lin2023directional}
Ju Lin, Niko Moritz, Ruiming Xie, Kaustubh Kalgaonkar, Christian Fuegen, and
  Frank Seide.
\newblock Directional speech recognition for speaker disambiguation and
  cross-talk suppression.
\newblock In \emph{Proc. Interspeech}, pages 3522--3526, 2023.

\bibitem[Lv et~al.(2024)Lv, Charron, Moulon, Gamino, Peng, Sweeney, Miller,
  Tang, Meissner, Dong, et~al.]{lv2024aria}
Zhaoyang Lv, Nicholas Charron, Pierre Moulon, Alexander Gamino, Cheng Peng,
  Chris Sweeney, Edward Miller, Huixuan Tang, Jeff Meissner, Jing Dong, et~al.
\newblock Aria everyday activities dataset.
\newblock \emph{arXiv preprint arXiv:2402.13349}, 2024.

\bibitem[Ma et~al.(2024)Ma, Ye, Hong, Guzov, Jiang, Postyeni, Pesqueira,
  Gamino, Baiyya, Kim, et~al.]{ma2024nymeria}
Lingni Ma, Yuting Ye, Fangzhou Hong, Vladimir Guzov, Yifeng Jiang, Rowan
  Postyeni, Luis Pesqueira, Alexander Gamino, Vijay Baiyya, Hyo~Jin Kim, et~al.
\newblock Nymeria: A massive collection of multimodal egocentric daily motion
  in the wild.
\newblock In \emph{European Conference on Computer Vision}, pages 445--465.
  Springer, 2024.

\bibitem[Pan et~al.(2023)Pan, Charron, Yang, Peters, Whelan, Kong, Parkhi,
  Newcombe, and Ren]{pan2023aria}
Xiaqing Pan, Nicholas Charron, Yongqian Yang, Scott Peters, Thomas Whelan, Chen
  Kong, Omkar Parkhi, Richard Newcombe, and Yuheng~Carl Ren.
\newblock Aria digital twin: A new benchmark dataset for egocentric 3d machine
  perception.
\newblock In \emph{Proceedings of the IEEE/CVF International Conference on
  Computer Vision}, pages 20133--20143, 2023.

\bibitem[Perrett et~al.(2025)Perrett, Darkhalil, Sinha, Emara, Pollard, Parida,
  Liu, Gatti, Bansal, Flanagan, et~al.]{perrett2025hd}
Toby Perrett, Ahmad Darkhalil, Saptarshi Sinha, Omar Emara, Sam Pollard,
  Kranti~Kumar Parida, Kaiting Liu, Prajwal Gatti, Siddhant Bansal, Kevin
  Flanagan, et~al.
\newblock Hd-epic: A highly-detailed egocentric video dataset.
\newblock In \emph{Proceedings of the Computer Vision and Pattern Recognition
  Conference}, pages 23901--23913, 2025.

\bibitem[Straub et~al.(2024)Straub, DeTone, Shen, Yang, Sweeney, and
  Newcombe]{straub2024efm3d}
Julian Straub, Daniel DeTone, Tianwei Shen, Nan Yang, Chris Sweeney, and
  Richard Newcombe.
\newblock Efm3d: A benchmark for measuring progress towards 3d egocentric
  foundation models.
\newblock \emph{arXiv preprint arXiv:2406.10224}, 2024.

\bibitem[Team(2025)]{AriaGen2Device}
Project~Aria Team.
\newblock Aria gen 2: An advanced research device for egocentric ai research,
  2025.
\newblock www.projectaria.com/ariagen2devicepaper.

\bibitem[Wen et~al.(2025)Wen, Trepte, Aribido, Kautz, Gallo, and
  Birchfield]{wen2025foundationstereo}
Bowen Wen, Matthew Trepte, Joseph Aribido, Jan Kautz, Orazio Gallo, and Stan
  Birchfield.
\newblock Foundationstereo: Zero-shot stereo matching.
\newblock In \emph{Proceedings of the Computer Vision and Pattern Recognition
  Conference}, pages 5249--5260, 2025.

\bibitem[Yang et~al.(2025)Yang, Liu, Guo, Dong, Zhang, Zhang, Wang, Zhou, Xie,
  Wang, et~al.]{yang2025egolife}
Jingkang Yang, Shuai Liu, Hongming Guo, Yuhao Dong, Xiamengwei Zhang, Sicheng
  Zhang, Pengyun Wang, Zitang Zhou, Binzhu Xie, Ziyue Wang, et~al.
\newblock Egolife: Towards egocentric life assistant.
\newblock In \emph{Proceedings of the Computer Vision and Pattern Recognition
  Conference}, pages 28885--28900, 2025.

\end{thebibliography}
